\documentclass[11pt,a4paper]{article}
\usepackage[hyperref]{emnlp2020}
\usepackage{times}
\usepackage{latexsym}

\usepackage{comment}
\usepackage{graphicx}
\usepackage{xcolor,colortbl}

\usepackage{microtype}

\aclfinalcopy % Uncomment this line for the final submission
\newcommand{\isep}{\mathrel{{.}\,{.}}\nobreak}

\title{\textsc{Artemis}: A Novel Annotation Methodology for Indicative Single Document Summarization}

\author{\\{\bf Rahul Jha}$^\star$, {\bf Keping Bi}$^\dag$, {\bf Yang Li}$^{\star}$,  {\bf Mahdi Pakdaman}$^{\star}$ \\ {\bf Asli Celikyilmaz}$^\star$, {\bf Ivan Zhiboedov}$^\ddag$, {\bf Kieran McDonald}$^{\star}$ \\ 
$^\star$ Microsoft Corporation\\
$^\dag$ Umass Amherst\\
$^\ddag$ Facebook Inc}

\date{}

\begin{document}
\maketitle
\begin{abstract}
We describe \textsc{Artemis} (Annotation methodology for Rich, Tractable, Extractive, Multi-domain, Indicative Summarization), a novel hierarchical annotation process that produces indicative summaries for documents from multiple domains. Current summarization evaluation datasets are single-domain and focused on a few domains for which naturally occurring summaries can be easily found, such as news and scientific articles. These are not sufficient for training and evaluation of summarization models for use in document management and information retrieval systems, which need to deal with documents from multiple domains. Compared to other annotation methods such as Relative Utility and Pyramid, \textsc{Artemis} is more tractable because judges don't need to look at all the sentences in a document when making an importance judgment for one of the sentences, while providing similarly rich sentence importance annotations. We describe the annotation process in detail and compare it with other similar evaluation systems. We also present analysis and experimental results over a sample set of 532 annotated documents.
\end{abstract}

\section{Introduction}
\label{sec:intro}
\footnotetext[2]{Work done while an intern at Microsoft.}
\footnotetext[3]{Work done while an employee of Microsoft.}
\begin{figure}\small
% Doc Id: 46214.docx from dev data
    \centering
    \begin{tabular}{|p{7.3cm}|}
    \hline
    \multicolumn{1}{|c|}{\textbf{Original Document}} \\
    \hline
    \textit{(1)} This content should be viewed as reference documentation only, to inform IT business decisions \ldots \\
    \textit{(2)} Microsoft employees need to stay aware of new company products, services, processes, and personnel-related developments in an organization that provides them \ldots\\
    \textit{(3)} The SMSG Readiness team at Microsoft developed a suite of applications that delivers training and information to Microsoft employees according to employee roles \ldots\\
    \textit{(4)} Microsoft Information Technology (Microsoft IT) is responsible for managing one of the largest Information Technology (IT) infrastructure environments in the world.\\
    \textit{(5)} It consists of 95,000 employees working in 107 countries worldwide.\\
    \textit{(6)} The Sales, Marketing, and Services Group (SMSG) at Microsoft is responsible for servicing the needs of Microsoft customers and partners.\\
    \textit{(7)} It is essential that these 45,000 employees remain informed about products and services within their areas of expertise and, in turn, to educate and inform \ldots\\
    \textit{(8)} The SMSG Readiness (SMSGR) team at Microsoft is responsible for ensuring that SMSG employees have all of the tools and knowledge they require to deliver \ldots \\
    \textit{(\ldots document truncated)} \\
        \hline
    \multicolumn{1}{|c|}{\textbf{Summary 1}} \\
    \hline
    \textit{(2)} Microsoft employees need to stay aware of new company products, services, processes, and \ldots \\
    \textit{(3)} The SMSG Readiness team at Microsoft developed a suite of applications that delivers training and \ldots\\
    \textit{(4)} Microsoft Information Technology (Microsoft IT) is responsible for managing one of the largest \ldots\\
    \hline
    \multicolumn{1}{|c|}{\textbf{Summary 2}} \\
    \hline
    \textit{(3)} The SMSG Readiness team at Microsoft developed a suite of applications that delivers training and \ldots\\
    \textit{(6)} The Sales, Marketing, and Services Group (SMSG) at Microsoft is responsible for servicing the needs of \ldots\\
    \textit{(8)} The SMSG Readiness (SMSGR) team at Microsoft is responsible for ensuring that SMSG employees have \ldots \\
    \hline
    \multicolumn{1}{|c|}{\textbf{Summary 3}} \\
    \hline
    \textit{(2)} Microsoft employees need to stay aware of new company products, services, processes, and \ldots \\
    \textit{(4)} Microsoft Information Technology (Microsoft IT) is responsible for managing one of the largest \ldots\\
    \textit{(8)} The SMSG Readiness (SMSGR) team at Microsoft is responsible for ensuring that SMSG employees have \ldots\\
    \hline
    \end{tabular}
    \caption{One of the documents from our sample annotated dataset along with indicative summaries annotated by three different judges. The sentence numbers in round brackets are not in the original document but are added here for readability. Summary sentences are truncated for readability as well.}
    \label{fig:annotation_example}
\end{figure}

Given an input source document, summarization systems produce a condensed summary which can be either informative or indicative. Informative summaries try to convey all the important points of the document \cite{Kan02usingthe,Kan01domain-specificinformative}, while indicative summaries hint at the topics of the document, pointing to information alerting the reader about the document content \cite{saggion&alapalme}. 
An informative summary aims to replace the source document, so that the user does not need to read the full document \citep{Edmundson:1969:NMA:321510.321519}. An indicative summary, on the other hand, aims to help the user decide whether they should consider reading the full document \cite{Kan02usingthe}. 

The content of indicative summaries can be composed in several ways. For example, it can contain sentences extracted from the source document which relate to its main topic \cite{Barzilay97usinglexical,kupiec95}, generated text describing how a document is different from other documents \cite{Kan01applyingnatural}, topic keywords \cite{Hovy1997AutomatedTS,saggion&alapalme} as well as metadata such as length and writing style \cite{KathyAni}.

Document management systems such as Microsoft OneDrive and SharePoint, Google Docs and Dropbox can use indicative summaries to help their users decide whether a given document is relevant for them before opening the full document. Indicative summaries can also be used in information retrieval systems as previews for documents returned in search results. Document summarization systems deployed in these real-world systems need to be able to summarize documents from a wide variety of domains. 

However, existing summarization datasets are highly domain-specific, with a majority of them focusing on news summarization \citep{nallapati-etal-2016-abstractive, grusky18, nyt2008,graff2003english}. One of the reasons for this bias towards news summarization is the availability of naturally occurring summaries for news, which makes it easier to create large-scale summarization datasets automatically by scraping online sources. Apart from the domain bias, they are also susceptible to noise which can affect upto 5.92\% of the data \citep{Kryscinski2019NeuralTS}.

\begin{figure*}
  \centering
\includegraphics[width=0.95\textwidth]{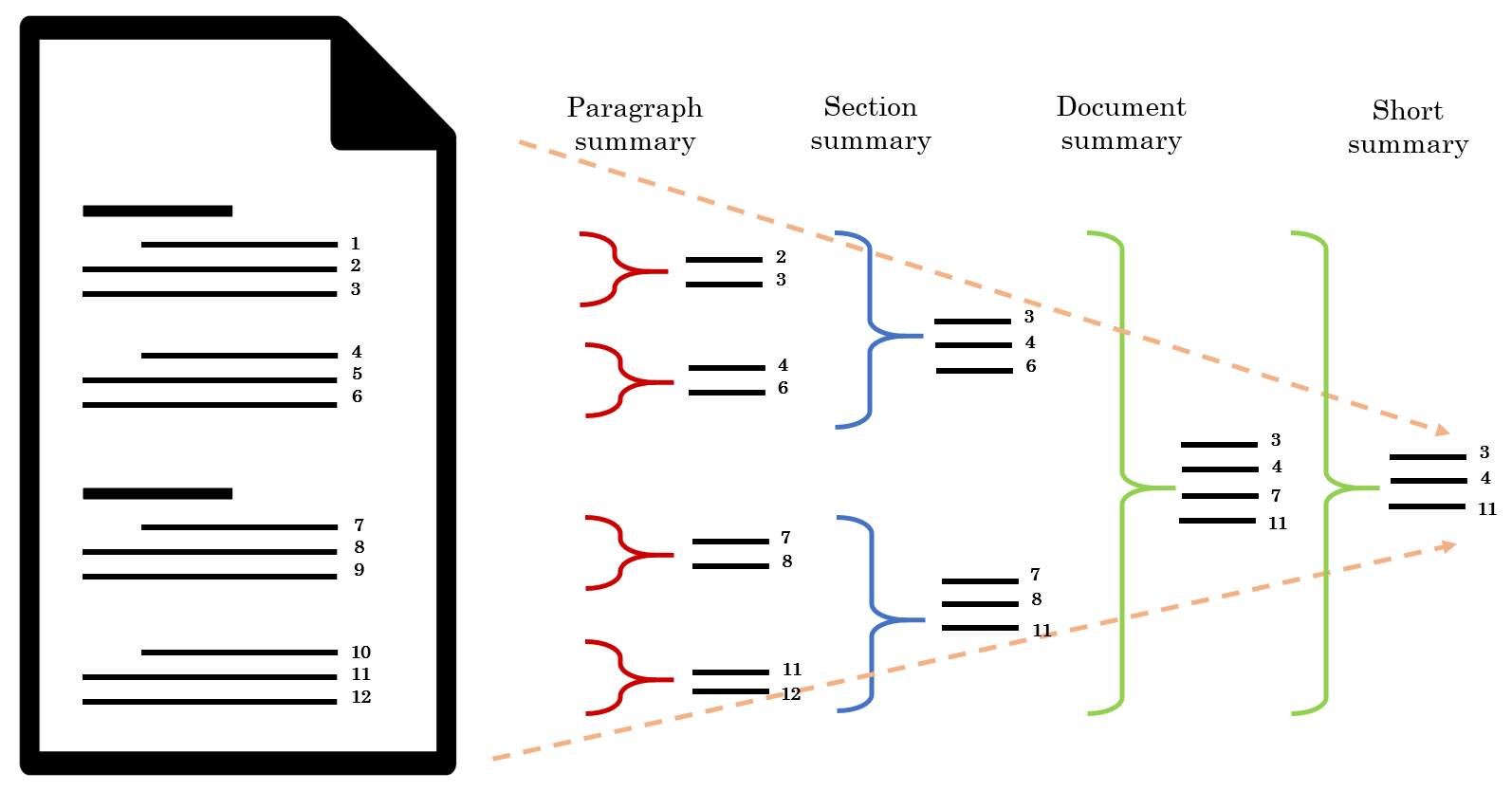}
\caption{A schematic of \textsc{Artemis} annotation process. A document is divided into sections and paragraphs. The judges summarize paragraphs, sections and the document hierarchically, at each step using sentences selected at the previous step.} 
  \label{fig:annotation_process}
\end{figure*}

In order to train and evaluate multi-domain summarization models for use in document management systems, we need to build representative datasets geared towards this use case. Towards this goal, we present \textsc{Artemis} (Annotation methodology for Rich, Tractable, Extractive, Multi-domain, Indicative Summarization), a hierarchical annotation process for indicative summarization of multi-domain documents. Figure~\ref{fig:annotation_example} shows a sample document with three annotated summaries obtained using \textsc{Artemis}.

\textsc{Artemis}'s hierarchical annotation process allows judges to create indicative summaries for long documents through divide-and-conquer. Judges successively summarize larger and larger chunks of a document in multiple stages, at each stage reusing sentences selected previously. The hierarchical process means that judges only look at a small set of sentences at each stage.

Compared to previous annotation methods, where judges need to consider all the document sentences together when building a summary \cite{Radev&al.07b} or create expensive semantic annotations \cite{nenkova-passonneau-2004-pyramid}, \textsc{Artemis} is a low-cost annotation approach that produces rich sentence importance annotations. Judges are able to use \textsc{Artemis} to annotate documents averaging 1322 words (77 sentences) in 4.17 minutes on average, based on an initial sample of annotation tasks. This is almost twice the length of summarization datasets such as CNN/Dailymail at 766 words \cite{nallapati-etal-2016-abstractive} and \textsc{Newsroom} at 659 words \cite{grusky18}.

\textsc{Artemis}'s annotation process aims at selecting a set of sentences that contain relevant information about the main topics of a document rather than conveying all the relevant information in a document. Given this, summaries annotated by \textsc{Artemis} are indicative in nature and suited for document management and information retrieval systems, where they can be used as part of document preview to help a user decide whether a document is relevant for them.

The rest of this paper is organized as follows. Section~\ref{sec:method} describes the annotation process in detail using illustrative examples. Section~\ref{sec:rel} relates our method to previous annotation methods for summarization. Section~\ref{sec:analysis} presents a number of analyses characterizing the \textsc{Artemis} annotation process in terms of label distribution and judge agreement by using a sample annotated document set. Section~\ref{sec:experiments} presents evaluation results for a set of baseline summarization models on the sample annotated document set. We present some final concluding remarks in Section~\ref{sec:conclusion}.

\section{Annotation Methodology}
\label{sec:method}
Figure~\ref{fig:annotation_process} shows a high-level diagram representing the annotation process for \textsc{Artemis}. Given a document as input, the preprocessing step consists of first dividing the document into sections, each of which is further divided into paragraphs. The section and paragraph boundaries are computed based on a set of heuristics that depend on signals like explicit section headers as well as constraints on the number of sentences shown at each screen. 

The hypothetical document in Figure~\ref{fig:annotation_process} is divided into two sections with two paragraphs each. The first section contains sentences $\{1\isep6\}$, with two paragraphs containing sentences $\{1\isep3\}$ and $\{4\isep6\}$ respectively. The second section contains sentences $\{7\isep12\}$, again with two paragraphs containing sentences $\{7\isep9\}$ and $\{10\isep12\}$. 

A salient sentence is defined as a sentence that includes a main concept or idea for summarizing the text, or a fact or an argument emphasized by the author\footnote{Authors can emphasize sentences either through formatting or discourse cues.}. Several example sentences are provided in the judge guidelines to help them distinguish salient sentences from non-salient sentences. At a high-level, the judges are trained to select sentences that allow a reader to decide whether to read the full document or not.

To summarize the document, judges proceed in a bottom-up manner starting from paragraphs (left-to-right in Figure~\ref{fig:annotation_process}). A judge is first asked to summarize each paragraph in a section by selecting a few salient sentences. A minimum number of sentences are required for each paragraph-summary \footnote{The judge can also mark incomplete and grammatically incorrect sentences as defective, which are not counted when computing the minimum threshold for paragraph-summary.}. Once a paragraph has been summarized, the annotation continues to the next paragraph till paragraph-level summaries are created for all the paragraphs in a section. For the document in Figure~\ref{fig:annotation_process}, the judge selected sentences $\{2,3\}$ for the first paragraph and sentences $\{4,6\}$ for the second paragraph.

Once all the paragraphs in a section are summarized, the judge is asked to create a summary for the entire section. However, the judge doesn't have to look at all the sentences in the section to build the section-level summary. Instead, they only select from the set of sentences previously selected to summarize the paragraphs of the section. For example, for summarizing the first section in Figure~\ref{fig:annotation_process}, the judge only needs to select from the set of sentences $\{2,3,4,6\}$, instead of the entire set of sentences $\{1\isep6\}$ that comprise the section. In the example, the judge decided to use the sentences $\{3,4,6\}$ for summarizing the first section.

\begin{table*}\small
% Doc Id: 95619.docx from dev data
    \centering
    \begin{tabular}{|p{10.8cm}|c|c|c|c|}
    \hline
    \textbf{Document Sentences} & \textbf{\#Para} & \textbf{\#Sec} & \textbf{\#Doc} & \textbf{\#Short} \\
    \hline
    \textit{(1)} This content should be viewed as reference documentation only, to inform IT \ldots & 0 & 0 & 0 & 0\\
    \hline
    \textit{(2)} Microsoft employees need to stay aware of new company products, services, processes, and personnel-related developments in an organization that provides \ldots & 2 & 2 & 2 & 2\\
    \hline
    \textit{(3)} The SMSG Readiness team at Microsoft developed a suite of applications that delivers training and information to Microsoft employees according to employee \ldots & 4 & 4 & 4 & 3\\
    \hline
    \textit{(4)} Microsoft Information Technology (Microsoft IT) is responsible for managing one of the largest Information Technology (IT) infrastructure environments in the world. & 5 & 4 & 2 & 2\\
    \hline
    \textit{(5)} It consists of 95,000 employees working in 107 countries worldwide. & 0 & 0 & 0 & 0\\
    \hline
    \textit{(6)} The Sales, Marketing, and Services Group (SMSG) at Microsoft is responsible for servicing the needs of Microsoft customers and partners. & 3 & 1 & 1 & 1\\
    \hline
    \textit{(7)} It is essential that these 45,000 employees remain informed about products and services within their areas of expertise and, in turn, to educate and inform \ldots & 1 & 0 & 0 & 0\\
    \hline
    \textit{(8)} The SMSG Readiness (SMSGR) team at Microsoft is responsible for ensuring that SMSG employees have all of the tools and knowledge they require to deliver \ldots & 4 & 3 & 2 & 2 \\
    \hline
    \end{tabular}
    \caption{Detailed view of the annotation for the document shown in Figure~\ref{fig:annotation_example}. Against each sentence, we show the number of judges that selected the sentence at paragraph, section, document and short summary stage.}
    \label{tab:judge_counts}
\end{table*}

Once a section is summarized, the annotation proceeds to the next section in a similar manner. Once all the sections of a document are summarized, the judge is asked to build the document summary by selecting from sentences that they had previously selected to build the section-level summaries. In Figure~\ref{fig:annotation_process}, the judge selected sentences $\{3,4,7,11\}$ for the document level summary. Finally, the judge is asked to build a short summary for the document by selecting three most salient sentences from their document-level summary. 

\textsc{Artemis}'s hierarchical annotation process considerably reduces the cognitive load on the judges. By reusing judgements made at previous steps, judges are able to successively summarize long documents by divide-and-conquer. For creating the document-level summary in the hypothetical example in Figure~\ref{fig:annotation_process}, the judge only needs to look at the 6 sentences $\{3,4,6,7,8,11\}$ selected for the two section-level summaries, instead of having to go over the entire set of 12 sentences.

Table~\ref{tab:judge_counts} shows a more detailed view of the annotation for an actual document annotated through \textsc{Artemis} with five judges (This is the same document that was used in Figure~\ref{fig:annotation_example}). For each of the first eight sentences in the document, it shows the number of judges that selected the sentence at paragraph, section, document and short summary stage. This table gives an insight into the kind of information available from the annotation. 

Sentences \textit{(1)} and \textit{(5)} were deemed by every judge as not salient. Sentence \textit{(3)} was selected by four judges as salient up to document-summary level, but one of the judges dropped it at short-summary level. Similarly, sentence \textit{(4)} was selected at paragraph-summary level by five judges, but only two judges kept it till the document and short-summary level. In Section~\ref{sec:analysis}, we present statistics on a sample annotated document set that characterize the annotation process in more detail.

\section{Related Work}
\label{sec:rel}
We now compare \textsc{Artemis} with existing summarization evaluation methods. We start with discussing Relative Utility, which is most related to our methodology, and describe how \textsc{Artemis} obtains similar judgments, but with a light-weight process where judges don't need to look at the entire input document when annotating a sentence. Following this, we discuss DUC evaluations, ROUGE and the Pyramid method. Finally, we discuss some of the recent trends in summarization evaluation.
\subsection{Relative Utility}
\citet{Radev&al.07b} introduce Relative Utility (RU) as an evaluation metric to account for Summary Sentence Substitutability (SSS) problem in co-selection metrics. Co-selection metrics are evaluation metrics for extractive summarization that depend on text unit overlap with ideal reference summaries created by judges. The SSS problem arises because the judges only provide information about the sentences that they selected for a fixed-length summary. However, other sentences in the document might be equally good candidates for the summary. Human judges often disagree about which are the top n\% of the sentences in a document \cite{Mani01summarizationevaluation}. 

To address the SSS problem, in RU evaluation judges are asked to assign a utility score to each sentence in a document on a scale of 0 to 10. Given these utility scores, the score for any arbitrary extractive summary can be computed based on the utility of the sentences in the summary. 

In RU, to assign the utility score to a sentence in the document, a judge needs to compare the sentence with every other sentence in the document. This can be difficult for long documents. \textsc{Artemis} is a light-weight process that achieves an approximation of this. By assigning graded importance scores to paragraph, section, document and short summary level labels, we can obtain an approximate utility score for each sentence. For example scores $\{1,2,3,4\}$ could be assigned to sentences selected at paragraph, section, document and short summary level and a score of $0$ could be assigned to sentences not selected at any level.

\subsection{DUC evaluations and ROUGE} 
DUC (Document Understanding Conferences) were a series of conferences run to further progress in summarization. DUC 2001-2004 focused on single and multi-document summarization \cite{Dang05overviewof}. In DUC evaluation for summary content, first a single human judge creates a model summary for each document. The model summary is split automatically into content units. For evaluating a system generated summary, a human judge compares the sentences in the system summary with model content units and estimates the fact overlap. 

The use of a single model summary in DUC evaluations raised concerns in the research community and led to the proposal of Pyramid evaluation, which we describe in Section~\ref{sec:eval_pyramid}. \citet{lin2004looking} concluded that given enough samples, the use of single model summaries was valid, but using multiple model summaries increased correlation with human judgments.

In later years, DUC experimented with ROUGE \cite{lin-2004-rouge}, an automatic metric for summary evaluation that uses n-gram co-occurrence statistics for scoring system generated summaries against the model summaries. ROUGE is the standard automatic evaluation method used in recent summarization evaluations, which we describe in Section~\ref{recent_eval}. 

In \textsc{Artemis}, the sentences selected by judges for document or short-level summary can be used as model summaries for ROUGE evaluation, as we demonstrate in Section~\ref{sec:experiments}. In addition, the labels for sentences at different summary levels could be used to train a pair-wise sentence ranking system such as LambdaMart \cite{burges2010from} or come up with more refined evaluation metrics.

\subsection{Pyramid evaluation}
\label{sec:eval_pyramid}
\citet{nenkova-passonneau-2004-pyramid} introduced Pyramid method as a more reliable method for summary evaluation by incorporating the idea that no single best model summary exists. Given a set of human-generated model summaries for a document, the Pyramid method starts by manually identifying Summary Content Units (SCUs) in the model summaries. A SCU represents a single unit of information (e.g. ``Two men were indicted") which can have different surface realizations in different summaries (e.g. ``Court indicted two men", ``Two men have been indicted").

The weight of an SCU is the number of model summaries it appears in. Thus, an SCU appearing in five model summaries has a higher weight than an SCU appearing in three model summaries. Given the SCU inventory over all model summaries, the Pyramid score of a system generated summary is obtained based on the number and weights of the SCUs in the summary. \citet{nenkova-passonneau-2004-pyramid} observe that the number of SCUs grows as the number of model summaries increases, confirming a similar observation by \citet{van-halteren-teufel-2003-examining}, supporting the claim that different judges deem different facts as important.

Finding SCUs in model summaries and then matching them to system summaries is an expensive semantic judgment task. Once created, the SCU inventory can be used to assign an importance weight to any sentence in a system generated extractive summary based on the weights of SCUs in it. Our methodology provides a cheaper method for assigning importance weight for each sentence in a document. In \textsc{Artemis}, multiple judges select each sentence for multiple summaries at paragraph, section, document, and short-summary levels. These judgments provide a low-cost way of obtaining an importance weight for a sentence, without expensive SCU annotation.

\subsection{Recent Trends in Summarization Evaluation}
\label{recent_eval}
Recent summarization evaluations are done using large scale datasets collected automatically from the web. Most of these datasets are from the news domain, including CNN/DailyMail \citep{nallapati-etal-2016-abstractive}, \textsc{Newsroom} \cite{grusky18}, New York Times \cite{nyt2008} and Gigaword \cite{rush-etal-2015-neural}. Some of the other domains investigated are scientific articles \cite{cohan-etal-2018-discourse}, patents \cite{sharma-etal-2019-bigpatent}, and Reddit stories \cite{kim-etal-2019-abstractive}. 

Datasets built from naturally occurring summaries found online tend to focus on domains for which manually written summaries are easily available such as news and scientific articles. These datasets are not sufficient for building a multi-domain document summarization application. Additionally, given the nature of data collection, often only a single summary is available for each document. This makes error analysis of individual examples difficult because different judges might deem different information as summary-worthy \cite{louis-nenkova-2013-automatically} as discussed in Section~\ref{sec:eval_pyramid}. \textsc{Artemis} provides a methodology for obtaining rich summary annotations for open-domain documents with multiple judges.

Summaries collected from online sources are also prone to noise. \citet{Kryscinski2019NeuralTS} manually inspected CNN/DailyMail and \textsc{Newsroom} datasets and found that the problem of noisy data affects upto 5.92\% of the summaries in different splits. Examples of noise they found include links to other articles and news sources, placeholder texts, unparsed HTML code, and non-informative passages in the reference summaries. In \textsc{Artemis}, such noisy text is excluded from annotation by explicit labeling of defective sentences. 

\citet{hardy-etal-2019-highres} proposed a new summarization evaluation approach called \textsc{HighRES}, which uses multiple judges to highlight salient information in original documents. Once the highlights are obtained, a system summary can be evaluated manually by asking judges to compare the system summary against highlights, or by a modified ROUGE evaluation that weighs n-grams by the number of times they were highlighted. \textsc{HighRES} is complementary to our hierarchical annotation approach and both the methods can be used together for obtaining rich summary annotations.

%Additionally, the summaries are often abstractive, and need heuristic matching to obtain extractive labels. Even with these labels, we cannot get the sort of rich annotation that are available in RU or Pyramid or Artemis.

\begin{table}\small
    \centering
    \begin{tabular}{|l|c|c|}
        \hline
        \textbf{Partition} & \textbf{\# Sentences} & \textbf{\# Documents} \\
        \hline
         Train & 19748 & 266 \\
         \hline
         Dev & 11488 & 138 \\
         \hline
         Test & 9898 & 128 \\
         \hline
    \end{tabular}
    \caption{Sample dataset used for the data analysis in this paper.}
    \label{tab:data_stats}
\end{table}

\section{Annotated Data Analysis}
\label{sec:analysis}

We present analysis on a sample dataset of 532  Microsoft Word documents crawled from the web with no domain restrictions. The data was annotated by a set of managed judges who were trained extensively for \textsc{Artemis} annotation process using detailed guidelines and illustrative examples. For additional quality control, we used a set of gold documents annotated by the development team for initial qualification tests for the judges as well as their ongoing evaluation. We divided the sample dataset into train, dev and test partitions, as shown in Table~\ref{tab:data_stats}. Unless otherwise stated, the statistics presented are computed over the dev partition. 

\subsection{Distribution Statistics}
\label{sec:dist_stats}
Table~\ref{tab:partition_stats} shows how the sentences of a document are divided across paragraphs and sections for the annotations. On an average, there are about 7 sections and 15 paragraphs in each document. The number of sentences in each section averages about 12, while the number of sentences in each paragraph averages about 5. Note that when summarizing a section, a judge has to look at much smaller number of sentences than 12, thanks to the hierarchical annotation process.

\begin{figure}
  \centering
\includegraphics[width=0.5\textwidth]{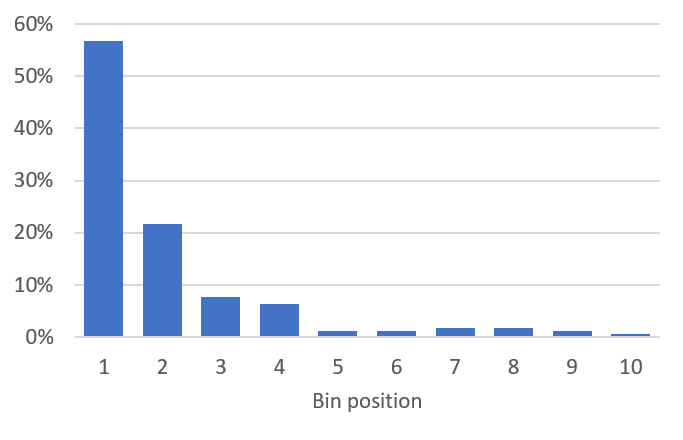} 
\caption{Distribution of sentences selected for doc-level summaries across 10 equally sized bins for each document.} 
  \label{fig:bin_distribution}
\end{figure}

To understand where the salient sentences lie for the documents, we divide each document into 10 equally sized bins and plot what fraction of sentences selected for the doc-level summaries lie in each bin. Each bin on an average contains $8.34\pm0.38$ sentences. Figure~\ref{fig:bin_distribution} shows the distribution of sentences selected for the doc-level summaries across the bins. More than 50\% of the selected sentences lie in the first bin and more than 90\% of the sentences lie in the first five bins. This shows that there is a bias for the summary sentences to be towards the first half of a document. However, the annotators don't form summaries by just selecting the first few sentences, as shown by the poor ROUGE-F1 scores obtained by the Lead-3 baseline in Section~\ref{sec:experiments}.

\begin{table}\small
    \centering
    \begin{tabular}{|l|p{2cm}|p{2.3cm}|}
        \hline
        \textbf{Partition} & \textbf{Average Count Per Document} & \textbf{Average number of sentences} \\
        \hline
        Section & 6.92 $\pm$ 0.91 & 12.04 $\pm$ 0.72 \\
        \hline
        Paragraph & 15.25 $\pm$ 2.17 & 5.46 $\pm$ 0.06 \\
        \hline

    \end{tabular}
    \caption{Average number of paragraphs and sections per document and the average number of sentences in each, along with the 95\% confidence interval.}
    \label{tab:partition_stats}
\end{table}

Another characterization of the annotation system can be done based on what fraction of salient sentences selected at each stage make it to the next stage. Table~\ref{tab:filtration_stats} shows this for all the stages of annotation. Looking at the diagonal first, we see that 82.44\% of the sentences selected as salient for paragraph-level summaries are also selected for section-level summaries, but only 69.57\% of the sentences selected for section-level summaries are selected for document-level summaries. From document-level summaries to short summary level, again 84.57\% of the salient sentences are kept. This shows that a larger number of sentences get filtered between the section and document level. Overall, only 48.51\% of the sentences selected for paragraph-level summaries are used for the final three-sentence short summaries.

\begin{table}\small
    \centering
    \begin{tabular}{|c|c|c|c|}
    \hline
    & \textbf{Section} & \textbf{Document} & \textbf{Short} \\
    \hline
    \textbf{Paragraph} & 82.44\% & 57.36\% & 48.51\% \\
    \hline
    \textbf{Section} & \cellcolor{lightgray} & 69.57\% & 58.84\%  \\
    \hline 
    \textbf{Document} & \cellcolor{lightgray} & \cellcolor{lightgray} & 84.57\%\\
    \hline
    \end{tabular}
    \caption{Filtration ratios for salient sentences between different stages. For example, the first row (Paragraph) shows what percentage of sentences selected at paragraph level survive till section, document and short-summary level. Table cells corresponding to filtration between same or out-of-order stages in the pipeline are colored gray.}
    \label{tab:filtration_stats}
\end{table}

\subsection{Agreement Statistics}
We compute Krippendorff's alpha over the entire annotated document set by treating each of paragraph, section, document and short summary level judgements as ordinal ratings. Across the set of all judges, the Krippendorff's alpha is 0.46. This is consistent with previous findings that summary content selection is a subjective task with moderate agreement \cite{Mani01summarizationevaluation}. 

For additional agreement evaluation, we had 10 documents evaluated by two sets of judges. The first set of judges was comprised of 4 developers involved in the design of \textsc{Artemis} and its guidelines. The second set of judges was comprised of 5 managed judges trained for doing the annotations. For each set of judges, a sentence was considered to be selected for document-level summary if at least 2 judges selected it. Given these judgements, the Kappa score between the two sets of judges was 0.43, which is considered moderate agreement \cite{landis_koch}.

Table~\ref{tab:agreement_stats} shows the average number of sentences selected at the different annotation levels if we use a minimum of 1, 2, or 3 judges to mark a sentence as salient out of the 5 total judges that annotate each document. We see that with 2 judges, there is agreement for 2.4 sentences for the final short summary, which is restricted to 3 sentences per judge. Even with 3 judges, there is agreement on 1.3 sentences for the final short-summary level.

\begin{table}\small
    \centering
    \begin{tabular}{|c|c|c|c|c|}
        \hline
         \textbf{\#} & \textbf{Paragraph} & \textbf{Section} & \textbf{Document} & \textbf{Short} \\
         \hline
        1 & 11.2 $\pm$ 1.0 & 9.4 $\pm$ 0.8 & 6.5 $\pm$ 0.4 & 5.4 $\pm$ 0.3 \\
        \hline
        2 & 5.1 $\pm$ 0.6 & 4.1 $\pm$ 0.5 & 2.8 $\pm$ 0.3 & 2.4 $\pm$ 0.2 \\
        \hline
        3 & 2.5 $\pm$ 0.4 & 2.0 $\pm$ 0.3 & 1.5 $\pm$ 0.2 & 1.3 $\pm$ 0.2 \\
         \hline
    \end{tabular}
    \caption{Average number of salient sentences at each stage corresponding to the minimum number of judges needed to mark a sentence as salient (out of a total of five judges) along with 95\% confidence intervals.}
    \label{tab:agreement_stats}
\end{table}

\section{Experiments}
\label{sec:experiments}
We evaluate a number of baseline methods on the sample annotated document set, partitioned into train, dev and test as described in Table~\ref{tab:data_stats}. 

For these experiments, the document-level summary created by each judge for a document is treated as an independent reference summary and we evaluate the candidate summary against all the reference summaries using ROUGE-F scores. 

\begin{itemize}

\item \textbf{Lead-3} baseline selects first three sentences of a document as the summary.
\item \textbf{Oracle} scores are obtained using a jack-knifed procedure. Reference summary from each judge is considered a predicted summary and evaluated against all the other reference summaries for the document. The Oracle ROUGE score is computed by averaging the scores for all judge summaries. 
\item \textbf{Cheng\&Lapata} \cite{cheng-lapata-2016-neural} is an encoder-decoder summarization model where each sentence is first encoded using a CNN (Convolutional Neural Network). These sentence level encodings are then passed through an RNN (Recurrent Neural Network) to create contextual encodings for each sentence. The encoding for the final sentence of the document is fed into the decoder, which uses another RNN with attention over input sentence encodings to predict the label for each sentence. At each decoding step, the decoder state also depends on the probability of the previous sentence being part of summary. 
\item \textbf{SummaRunner} \cite{nallapati2017summarunner} uses a hierarchical RNN to compute contextual encodings for each sentence in the input. These encodings are average pooled and passed through a non-linear transformation to create an encoding for the document. In a second pass, a logistic layer makes a binary decision for each sentence based on the sentence encodings, the document representation as well as factors modeling previously selected summary sentences and sentence position.
\item \textbf{Seq2SeqRNN} is a method introduced in \citet{kedzie-etal-2018-content} that uses an RNN to encode the input sentences. A separate RNN based decoder is used to transform each sentence into a query vector which attends to the encoder output. The attention weighted encoder output and the decoder GRU output are used together to predict the output label.
\end{itemize}

\begin{table}\small
    \centering
    \begin{tabular}{|l|c|c|c|} 
    \hline
    \textbf{Method} & \textbf{Rouge-1} & \textbf{Rouge-2} & \textbf{Rouge-L}\\
    \hline
        Lead-3 & 44.94 & 34.37 & 43.39 \\
          \hline
        Cheng \& Lapata & 60.21 & 49.81 & 58.62 \\
          \hline
        SummaRunner & 63.56 & 53.57 & 61.89 \\
        \hline
        Seq2SeqRNN & 63.89 & 54.22 & 62.36 \\
        \hline
        Oracle & 73.28 & 66.60 & 72.20 \\
        \hline
    \end{tabular}
    \caption{Results for different baselines on the test data.}
    \label{tab:results}
\end{table}

We used the code released by \citet{kedzie-etal-2018-content} for reproducing Cheng\&Lapata, SummaRunner and Seq2SeqRNN systems. The ROUGE-F score for each system on the test data is shown in  Table~\ref{tab:results}. 

The Lead baseline achieves a ROUGE-1 score of 44.94, which is significantly lower than the other systems as well as the Oracle. This shows that compared to news summarization, selecting the first few sentences is a much weaker baseline for open-domain summarization. 

The SummaRunner system does better than Cheng\&Lapata, potentially due to its incorporating multiple signals for content, salience, novelty and position. Seq2SeqRNN performs the best, which is consistent with the results reported in \citet{kedzie-etal-2018-content}. There is still a gap between these systems and the Oracle method, which achieves a ROUGE-1 score of 73.28.

\section{Concluding Remarks}
\label{sec:conclusion}
In this paper, we described \textsc{Artemis}, a novel hierarchical annotation methodology for indicative, extractive summarization. We described the annotation process in detail and compared it with Relative Utility, DUC evaluation methodology, the Pyramid method as well as other recent methods for summary content evaluation. We also presented analysis over a sample annotated dataset to characterize various properties of annotation process such as distribution of salient sentences and judge agreement. Finally, we showed experimental results for a set of baseline summarization systems using the annotated dataset.

Indicative summaries are useful in a number of scenarios involving information triage such as document management and information retrieval systems. However, summarization models for such systems need to be able to summarize documents from multiple domains. Most existing summarization datasets are single-domain and focused towards news, and hence are not sufficient for training and evaluating models for these applications. \textsc{Artemis} provides a low-cost methodology for annotating multi-domain indicative summaries compared to systems such as Pyramid and Relative Utility while producing similarly rich annotations.

\textsc{Artemis} summary annotations contain sentences that provide information about important topics in the document. The summaries are indicative because they do not aim to convey all the important points for a given information need, but instead, give a sense of what topics are covered in the document. The set of annotations in \textsc{Artemis} can be seen as a coarse partitioning between important and non-important sentences in an input document. Thus, models trained on these annotations can also be used as an importance signal in a larger pipeline for creating informative summaries.

\bibliographystyle{acl_natbib}
\bibliography{refs}

\end{document}